\documentclass[conference]{IEEEtran}

\IEEEoverridecommandlockouts
\usepackage{cite}
\usepackage{amsmath,amssymb,amsfonts}
\usepackage{algorithmic}
\usepackage{graphicx}
\usepackage{textcomp}
\usepackage{xcolor}
\usepackage{booktabs}
\usepackage{array}
\usepackage{multirow}
\usepackage{pifont}
\usepackage{balance}
\usepackage{url}
\def\BibTeX{{\rm B\kern-.05em{\sc i\kern-.025em b}\kern-.08em
    T\kern-.1667em\lower.7ex\hbox{E}\kern-.125emX}}

\begin{document}

\title{Guess What I Think: Streamlined EEG-to-Image Generation with Latent Diffusion Models\\
{\footnotesize }
\thanks{*Equal contribution.
%
%
This work was partially supported by the Italian Ministry of University and Research (MUR) within the PRIN 2022 Program for the project ``EXEGETE: Explainable Generative Deep Learning Methods for Medical Signal and Image Processing", under grant number 2022ENK9LS, CUP B53D23013030006, in part by the European Union under the National Plan for Complementary Investments to the Italian National Recovery and Resilience Plan (NRRP) with project PNC 0000001 D3 4 Health - SPOKE 1 - CUP B53C22006120001, and in part by Regione Lazio, Project “Deep Learning Generativo nel Dominio Ipercomplesso per Applicazioni di Intelligenza Artificiale ad Alta Efficienza Energetica”, under grant number 21027NP000000136.
}
}

\author{\IEEEauthorblockN{Eleonora Lopez\textsuperscript{*}, Luigi Sigillo\textsuperscript{*}, Federica Colonnese, Massimo Panella and Danilo Comminiello}
        \IEEEauthorblockN{\textit{Dept. Information Engineering, Electronics and Telecommunications (DIET), Sapienza University of Rome, Italy}\\
        Email: eleonora.lopez@uniroma1.it.}
}

\maketitle

\begin{abstract}

Generating images from brain waves is gaining increasing attention due to its potential to advance brain-computer interface (BCI) systems by understanding how brain signals encode visual cues. Most of the literature has focused on fMRI-to-Image tasks as fMRI is characterized by high spatial resolution. However, fMRI is an expensive neuroimaging modality and does not allow for real-time BCI. On the other hand, electroencephalography (EEG) is a low-cost, non-invasive, and portable neuroimaging technique, making it an attractive option for future real-time applications. Nevertheless, EEG presents inherent challenges due to its low spatial resolution and susceptibility to noise and artifacts, which makes generating images from EEG more difficult. In this paper, we address these problems with a streamlined framework based on the ControlNet adapter for conditioning a latent diffusion model (LDM) through EEG signals. We conduct experiments and ablation studies on popular benchmarks to demonstrate that the proposed method beats other state-of-the-art models. Unlike these methods, which often require extensive preprocessing, pretraining, different losses, and captioning models, our approach is efficient and straightforward, requiring only minimal preprocessing and a few components. The code is available at \url{https://github.com/LuigiSigillo/GWIT}.
    
\end{abstract}

\begin{IEEEkeywords}
EEG, Diffusion Models, Image Generation
\end{IEEEkeywords}

\section{Introduction}
\label{sec:introduction}
\begin{figure}
    \centering
    \includegraphics[width=0.98\columnwidth]{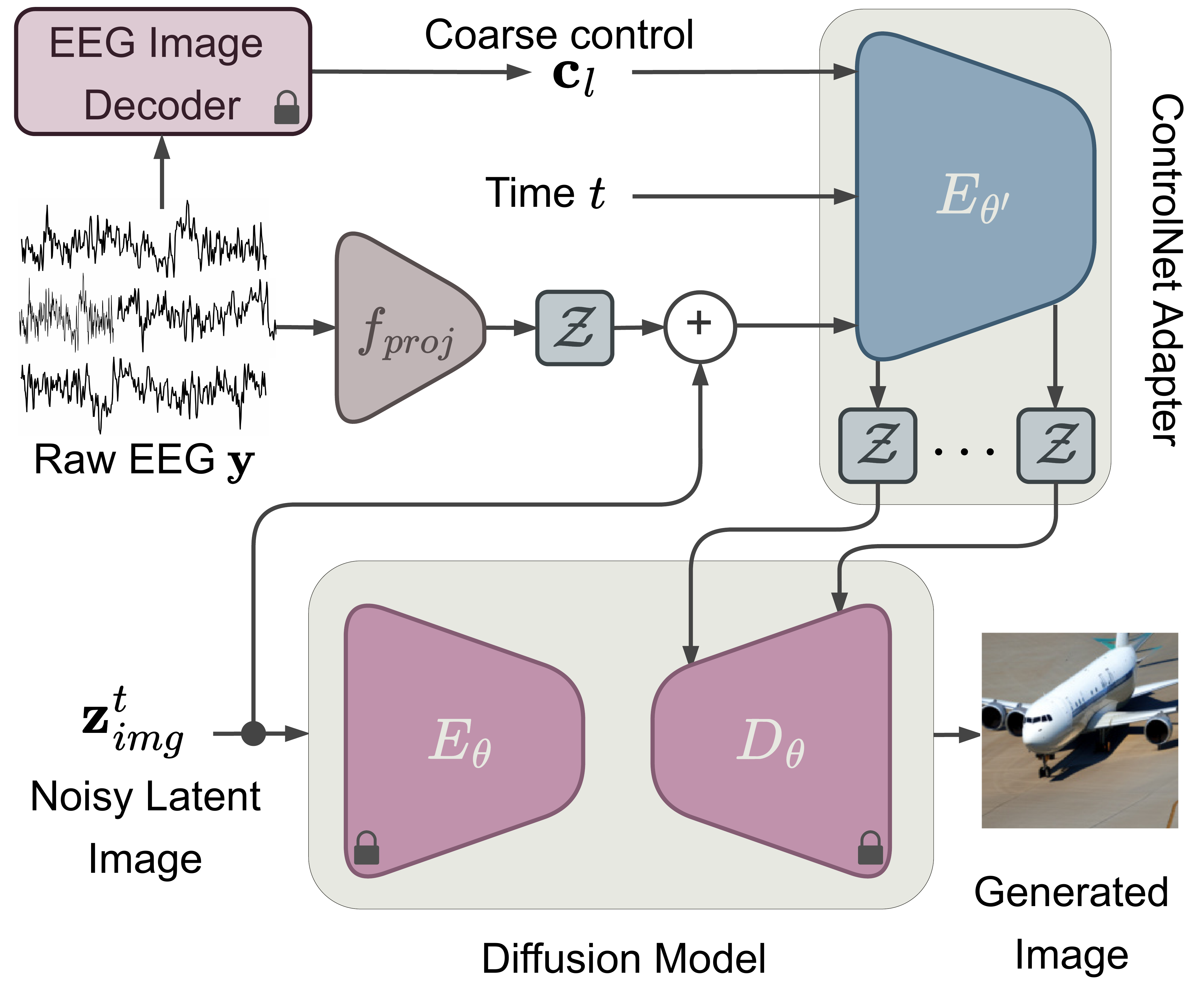}
    \caption{\textbf{Guess What I Think (GWIT).} Outline of our streamlined framework which includes a projection function, a ControlNet adapter to handle EEG conditioning, a frozen LDM, and a frozen EEG image decoder to obtain a coarse-grained control.}
    \label{fig:architecture}
\end{figure}

Advancing the Brain-Computer Interface (BCI) by understanding how the human brain represents the world is central to neurocognitive research. Indeed, BCIs have the potential to revolutionize areas such as healthcare, ranging from prevention to rehabilitation of neuronal injuries, as well as education and entertainment \cite{mudgal2020brain}. Among these advancements, some areas have been widely studied, such as emotion recognition \cite{lopez2023hypercomplex, lopez2024hierarchical, lopez2024phemonet} and neurodivergence classification \cite{stock2023towards, colonnese2025hyperdimensional}. More recently, with the advancement of generative models, the reconstruction of visual stimuli from brain signals, a task that had previously stagnated due to the limitations of earlier methods such as Generative Adversarial Networks (GANs), has resurfaced. 
\begin{figure*}[t]
    \centering
    \includegraphics[width=0.8\textwidth]{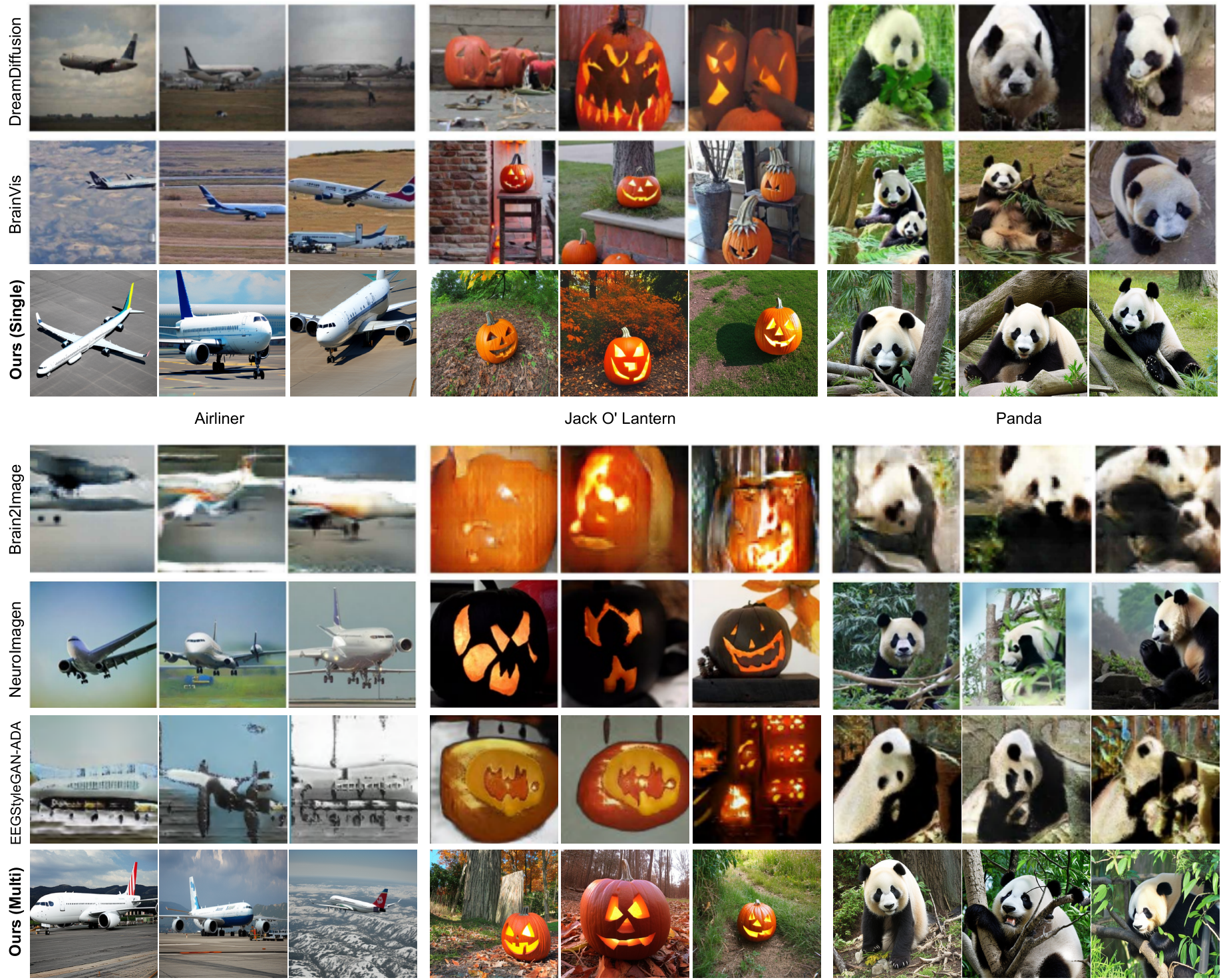}

    \caption{Comparison of images generated with models trained on subject $4$ (top three rows) and on all subjects (bottom four rows) of EEGCVPR40.}
    \label{fig:spampi_single}
\end{figure*}

Early neurocognitive works have shown that brain waves retain information about object structures from visual cues \cite{kourtzi2000cortical}. Building on this, many studies have since developed methods to reconstruct images from functional MRI (fMRI) signals \cite{chen2023seeing, zeng2024controllable, rombach2022high}. Thanks to its high spatial resolution and the generation abilities of diffusion models, these approaches are achieving increasingly accurate images. However, fMRI is cost-prohibitive because of the expensive equipment needed, and its lack of portability makes it unsuitable for real-time BCI systems \cite{lan2023seeing}. On the other hand, EEGs capture electrical activity in the brain through electrodes placed on the scalp, offering high temporal resolution. EEG is a portable, non-invasive, and low-cost neuroimaging technology, making it an appealing candidate for brain-to-image reconstruction and real-time applications \cite{mai2023brain}. Nonetheless, this task is inherently challenging, even with the high spatial resolution of fMRI data, and using EEG presents even more difficulties. In fact, EEGs are highly susceptible to noise, resulting in a very low signal-to-noise ratio (SNR), with artifacts frequently caused by factors like electrode misplacement or body movement \cite{mai2023brain}.


Studies addressing this task through EEG face three main drawbacks. First, they require extensive preprocessing that demands domain knowledge \cite{ferrante2024decoding}. Second, they rely on outdated generative models, such as GANs, which have been surpassed by more advanced diffusion models \cite{singh2023eeg2image, singh2024learning}. Third, these studies often employ complex frameworks that combine various alignment losses \cite{fu2023brainvis}, pertaining encoders on large datasets with different pretext tasks \cite{bai2023dreamdiffusion}, and captioning and silhouette extraction networks \cite{lan2023seeing}.

In this paper, we propose a streamlined framework, Guess What I Think (GWIT), that requires minimal preprocessing and utilizes ControlNet \cite{zhang2023adding} for adapting the EEG modality and for conditioning a latent diffusion model. Employing ControlNet has shown promising results for visual stimuli reconstruction from fMRI \cite{zeng2024controllable} and for music decoding from EEG \cite{postolache2024naturalistic}. To the best of our knowledge, this is the first time that ControlNet has been explored for image generation from EEG. Moreover, our approach requires minimal processing and is trained efficiently. We conduct thorough quantitative and qualitative evaluations on popular benchmarks surpassing state-of-the-art methods.

In Section~\ref{sec:related}, we review the related works. In Section~\ref{sec:method} we detail the proposed method for EEG-to-Image. Next, we describe the experimental setting and results in Section~\ref{sec:experiments} and draw conclusions in Section~\ref{sec:conclusion}.

\section{Related works}
\label{sec:related}

Advancements in generative methods have made it possible to reconstruct external stimuli, such as audio \cite{lee2023towards}, images \cite{takagi2023high}, and video \cite{chen2024cinematic}, from brain signal recordings. 
Given the high spatial resolution of fMRI data, numerous studies have investigated the use of fMRI to reconstruct visual stimuli \cite{chen2023seeing, takagi2023high, zeng2024controllable}. EEG, by contrast, is more accessible and economically viable than fMRI but it also presents more challenges, e.g., low SNR and spatial resolution. Many works exploring EEG-to-Image tasks are based on GANs \cite{kavasidis2017brain2image, tirupattur2018thoughtviz, singh2023eeg2image, singh2024learning}. In contrast, studies focused on reconstructing visual information from fMRI have shown highly promising results by leveraging diffusion models \cite{chen2023seeing, zeng2024controllable, takagi2023high, chen2024cinematic}. Indeed, diffusion models have achieved significant success across various tasks, including image generation \cite{rombach2022high}, super-resolution \cite{sigillo2024ship}, and audio generation \cite{comunita2024syncfusion}. Following this advancement, recent research has begun to develop diffusion-based methods for reconstructing images from EEG. Among these, many studies propose using alignment losses to mitigate the semantic gap between EEG and text/image data, developing EEG encoders that require pretraining on a large dataset \cite{fu2023brainvis, bai2023dreamdiffusion, li2024visual, ferrante2024decoding}, as well as integrating captioning models and silhouette extraction networks \cite{lan2023seeing, fu2023brainvis}. Moreover, many methods require extensive preprocessing that demands significant domain knowledge \cite{ferrante2024decoding, fu2023brainvis, bai2023dreamdiffusion}, whereas a recent approach has shown that excessive preprocessing can actually hinder image decoding performance \cite{singh2024learning}. 

\section{Proposed Method}
\label{sec:method}
\begin{figure}[t]
    \centering
    \includegraphics[width=0.8\linewidth]{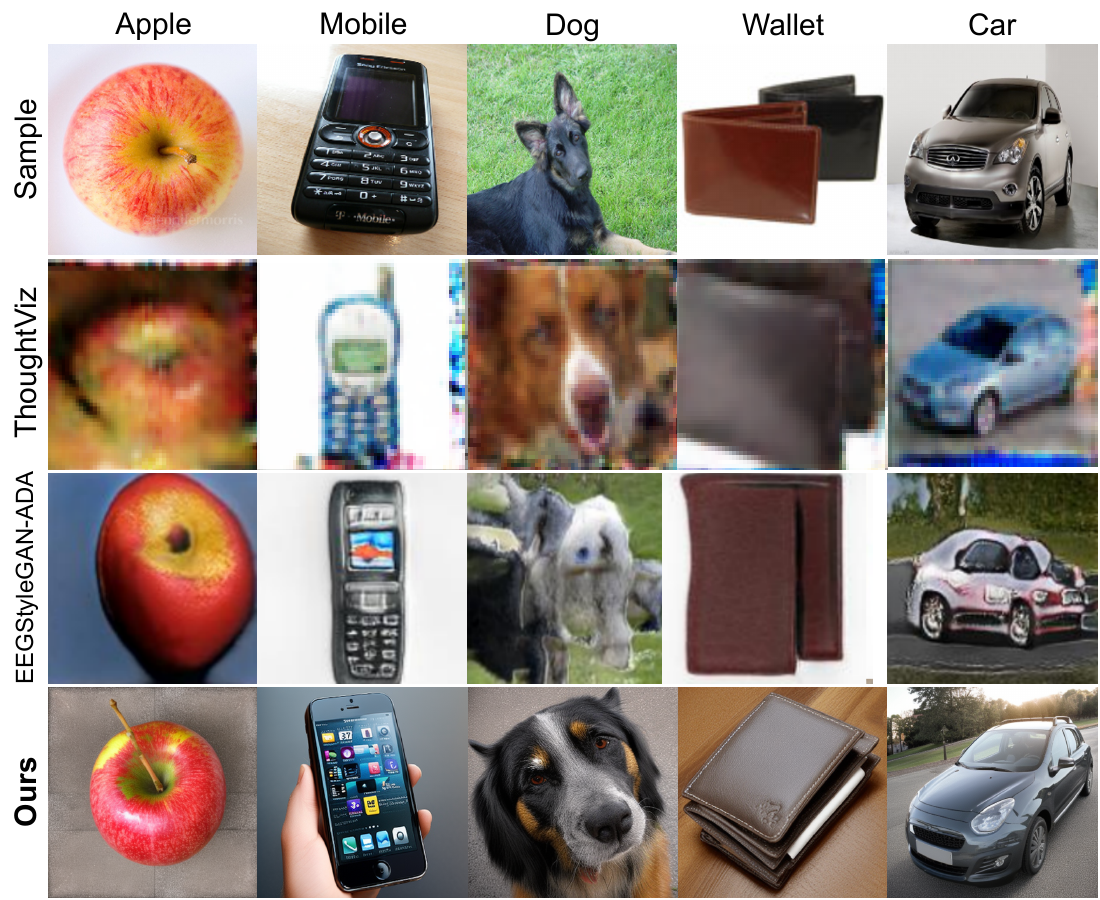}
    \caption{Comparison for images generated by models trained on ThoughtViz. First row: random sample from the dataset.}
    \label{fig:comparison_thoughtviz}
\end{figure}

We propose a simple framework, GWIT, comprising ControlNet as a mechanism for controlling a latent diffusion model \cite{rombach2022high} using EEG signals. 

\textbf{Problem formulation.} 
Let $\{\mathbf{y}, \mathbf{x}\}$ be a pair from the dataset, where $\mathbf{y} \in \mathbb{R}^{C \times L}$ is an EEG signal with $C$ channels and $L$ time steps, and $\mathbf{x} \in \mathbb{R}^{H_x \times W_x \times 3}$ is the corresponding image with height $H_x$, width $W_x$, and $3$ channels. The overall framework involves processing the image  $\mathbf{x}$ with the LDM in the standard manner, while the ControlNet adapter handles the input conditioning EEG $\mathbf{y}$, as it is able to adapt diverse modalities for conditioning
\cite{zhang2023adding, zeng2024controllable, postolache2024naturalistic}. Additionally, we add a coarse-grained control $\mathbf{c}_l$, i.e., a caption of the form ``Image of [label]", following the approach of \cite{ferrante2024decoding}. The label is predicted by a pretrained (frozen) EEG image decoder. The framework is then trained following the original ControlNet formulation, where the weights of the UNet encoder are copied and trained in an efficient way.

\textbf{Conditioning.}
The image is processed by the (pretrained) LDM, where it is mapped into a latent code $\mathbf{z}_{img} \sim E^{\text{VAE}}(\mathbf{\mathbf{x}})$ via the VAE stochastic encoder, with $\mathbf{z}_{img} \in \mathbb{R}^{H_z \times W_z \times D}$, where $H_z$ and $W_z$ are the spatial dimensions and $D$ the number of latent channels. Then, noise is progressively added through a forward Gaussian process indexed by $t \in [0, T]$ to obtain $\mathbf{z}_{img}^t$. The EEG is mapped into the same latent space of $\mathbf{z}_{img}$ via a simple $1$D convolutional neural network and reshape operations, i.e., $\mathbf{z}_{eeg} = f_{proj}(\mathbf{y})$, where $f_{proj}: \mathbb{R}^{C \times L} \rightarrow \mathbb{R}^{H \times W \times D}$. Then,  $\mathbf{z}_{eeg}$ is passed thorugh a $1 \times 1$ zero convolution layer and summed with $\mathbf{z}_{img}^t$, i.e., the control input to ControlNet is $\mathbf{c}_{eeg} = \mathbf{z}_{img}^t + \mathcal{Z}(\mathbf{z}_{eeg})$, with $\mathcal{Z}$ the zero-initilized convolution. We apply zero convolutions following the approach of ControlNet, as they prevent harmful noise from affecting the LDM backbone during the initial training steps \cite{zhang2023adding}. Then, the control input is processed by the ControlNet block. The adapter is defined as a trainable copy of the encoder of the underlying UNet architecture of the LDM which implements the diffusion model. That is, assuming the LDM with a UNet backbone formed by an encoder $E_\theta$ and a decoder $D_\theta$, the ControlNet adapter is defined as an encoder with a trainable copy of the weights $E_{\theta ^\prime}$. Thus, given input image $\mathbf{x}$, the corresponding noisy latent code $\mathbf{z}_{img}^t$, and a set of conditions including the time step $t$, the coarse control $\mathbf{c}_{l}$ and the EEG control $\mathbf{c}_{eeg}$, they are processed by the ControlNet $E_{\theta ^\prime}$ adapter as:
%
%
\begin{equation}
\begin{split}
\text{ControlNet}(\mathbf{c}_{eeg}, \mathbf{c}_l, t) &=
E_{\theta ^\prime}(\mathbf{c}_{eeg}, \mathbf{c}_l, t) \\
&= E_{\theta ^\prime} (\mathbf{z}_{img}^t + \mathcal{Z}(\mathbf{z}_{eeg}), \mathbf{c}_l, t) \\
&= E_{\theta ^\prime} (\mathbf{z}_{img}^t + \mathcal{Z}( f_{proj}(\mathbf{y})), \mathbf{c}_l, t).
\end{split}
\end{equation}

When training on multiple subjects we include a linear layer $S(\mathbf{y}, s)$ that encodes information on the subject by feeding it the subject id $s \in \mathbb{N}$, as done in \cite{defossez2023decoding, postolache2024naturalistic}. In this case, the EEG is processed as $f_{proj}(S(\mathbf{y}, s))$.

\textbf{Training.} The training loss used by ControlNet is identical to the original LDM loss, with two key differences: it includes an additional task-specific input condition, in this case, the EEG signal, and the UNet backbone remains frozen. Only the weights of the ControlNet adapter $E_{\theta ^\prime}$ and $f_{proj}$ are updated as follows:
\begin{equation}
    \mathcal{L} = \mathbb{E}_{\mathbf{z}_{img}^t, \mathbf{z}_{\text{eeg}}, \mathbf{c}_{l}, t, \epsilon \sim \mathcal{N}(0,1)} || \epsilon - \epsilon_\theta(\mathbf{z}_{img}^t, \mathbf{z}_{\text{eeg}}, \mathbf{c}_{l}, t)||_2^2,
\end{equation}

where $\epsilon_\theta$ is implemented by the UNet backbone and learns to predict the progressively added noise to the image. In this way, it can decode the image using the VAE-based decoder, i.e., the final generated image is $\hat{\mathbf{x}} = D^{\text{VAE}}(\hat{\mathbf{z}}_{img}^{0})$, where $\hat{\mathbf{z}}_{img}^{0}$ is the denoised sample of $\hat{\mathbf{z}}_{img}^{t} \sim \mathcal{N}(0,1)$.




 

\section{Experiments}
\label{sec:experiments}

\begin{table}[t] 
\centering 
\caption{Generation quality and semantic correctness of models trained on EEGCVPR40 (top: subject $4$, bottom: all subjects) and ThoughtViz.
}
\label{tab:bench_acc}
\resizebox{\linewidth}{!}{%
\begin{tabular}{llccc} 
\toprule
& \textbf{Model} & \textbf{IS} \textuparrow & \textbf{FID} \textdownarrow & \textbf{ACC} \textuparrow \\
\midrule
\multirow{10}{*}{\rotatebox{90}{\textbf{EEGCVPR40}}}
& DreamDiffusion \cite{bai2023dreamdiffusion} & \textemdash & \textemdash & $0.45$ \\
& BrainVis \cite{fu2023brainvis} & $31.52$ & $121.02$ & $0.49$ \\
& \textbf{GWIT (Ours)} &  $\mathbf{33.32}$ (\scalebox{0.8}{+6.34\%}) & $\mathbf{80.47}$ (\scalebox{0.8}{-33.51\%}) & $\mathbf{0.91}$ (\scalebox{0.8}{+85.71\%}) \\
\cmidrule{2-5}
& Brain2Image-GAN \cite{kavasidis2017brain2image} & $5.07$ & \textemdash & \textemdash\\ 
& NeuroVision \cite{khare2022neurovision} & $5.15$ & \textemdash & \textemdash \\
& Improved-SNGAN \cite{zheng2020decoding} & $5.53$ & \textemdash & \textemdash\\
& DCLS-GAN \cite{fares2020brain} & $5.64$ & \textemdash & \textemdash \\
& NeuroImagen \cite{lan2023seeing} & $33.50$ & \textemdash & $0.85$ \\
& EEGStyleGAN-ADA \cite{singh2024learning} & $10.82$ & $174.13$ & \textemdash \\
& \textbf{GWIT (Ours)} & $\mathbf{33.87}$ (\scalebox{0.8}{+1.1\%}) & $\mathbf{78.11}$ (\scalebox{0.8}{-55.14\%}) & $\mathbf{0.91}$ (\scalebox{0.8}{+7\%}) \\ 
\midrule
\multirow{6}{*}{\rotatebox{90}{\textbf{ThoughtViz}}} 
& AC-GAN \cite{odena2017conditional} & $4.93$ & \textemdash & \textemdash \\
& ThoughtViz \cite{tirupattur2018thoughtviz} & $5.43$ & \textemdash & \textemdash \\
& NeuroGAN \cite{mishra2023neurogan} & $6.02$ & \textemdash & \textemdash \\
& EEG2Image \cite{mustafa2012eeg} & $6.78$ & \textemdash & \textemdash \\
& EEGStyleGAN-ADA \cite{singh2024learning} &  $9.23$ & $109.49$ & \textemdash \\ 
& \textbf{GWIT (Ours)} & $\mathbf{13.76}$ (\scalebox{0.8}{+49.07\%}) & $\mathbf{66.33}$ (\scalebox{0.8}{-39.41\%}) & $\mathbf{0.787}$\\
\bottomrule
\end{tabular}
}
\vspace{-1mm}
\end{table}
%



\textbf{Datasets.}
We evaluate our method with the EEGCVPR40 \cite{spampinato2017deep, palazzo2020decoding} and ThoughtViz  \cite{kumar2018envisioned} benchmark datasets. The first contains EEG recordings from $6$ subjects who were shown $50$ images for each of $40$ classes from the ImageNet dataset \cite{deng2009imagenet}. EEGs were recorded for $0.5$s at $1$ kHz following the $128$-channel system. We used the official training, validation, and test sets. Instead, ThoughtViz contains recordings of EEG of $10$s sampled at $128$Hz with 14 electrodes from 23 participants. The samples were split into chunks of 32 time steps with overlap \cite{tirupattur2018thoughtviz}. We employed the subset of EEGs relative to images ranging in 10 classes. For preprocessing, we apply only standardization following \cite{singh2024learning} and directly employ raw EEG data.

\textbf{Implementation.}
For the projection $f_{proj}$ we use $1$D convolutional layers with (320, 640, 1280, 2560) channels and strides (5, 2, 2, 2). Finally, we apply padding and reshaping to map the EEG conditioning to the same dimension as the latent image. We employ Stable Diffusion \cite{rombach2022high} as LDM, using the weights of the $2.1$ version, and the LSTM EEG imade decoder proposed in \cite{singh2024learning}. 
The model, i.e., the ControlNet adapter and $f_{proj}$ are trained for $100$ epochs with Adam and a learning rate of $1e-5$. During training, we drop the coarse control for half of the samples to further emphasize the EEG conditioning \cite{zhang2023adding}. Lastly, for sampling the generated image we employ the guess mode of ControlNet which enforces the model to prioritize the ditioning over the coarse control \cite{zhang2023adding}. 

\begin{table}[t] 
\centering 
\caption{Ablation study for influence of EEG conditioning.}
\label{tab:ablation_eeg}
\begin{tabular}{lcc} 
\toprule
\textbf{Model} & \textbf{EEG Control} & \textbf{LPIPS} \textdownarrow \\
\midrule
GWIT-Only coarse & \ding{55} & $0.811$ \\
GWIT-Subject 4 & \ding{51} & $0.772$ \\
GWIT-All Subjects & \ding{51} & $0.770$ \\ 
\bottomrule
\end{tabular}
\end{table}

\begin{table}[t] 
\centering 
\caption{Ablation study for ``drop" (during training) and ``guess" (at inference time) modes on the multi-subject variant.}
\label{tab:ablation_drop_guess}
\begin{tabular}{lcccc} 
\toprule
\textbf{Model} & \textbf{Drop} & \textbf{Guess} & \textbf{LPIPS} \textdownarrow & \textbf{ACC} \textuparrow \\
\midrule
\multirow{4}{*}{GWIT}
 & \ding{55} & \ding{55} & $0.774$ & $0.59$\\
 & \ding{55} & \ding{51} & $0.770$ & $0.90$ \\
 & \ding{51} & \ding{55} & $0.781$ & $0.71$\\ 
 & \ding{51} & \ding{51} & $0.771$ & $0.91$\\ 
\bottomrule
\end{tabular}
\end{table}


\textbf{Metrics.}
We evaluate the generated images in terms of generation quality with Fréchet inception distance (FID) and inception score (IS). Moreover, we evaluate the semantic accuracy of generated images with N-way Top-k classification accuracy (ACC) \cite{chen2023seeing, lan2023seeing, fu2023brainvis}. This metric evaluates whether the original visual cue and the generated image are assigned to the same class by a pretrained ImageNet classifier, i.e., a ViT \cite{fu2023brainvis}. Finally, we employ the Learned Perceptual Image Patch Similarity (LPIPS) to measure the similarity between original and generated images.




%

\begin{figure}[t]
    \centering
    \includegraphics[width=0.8\linewidth]{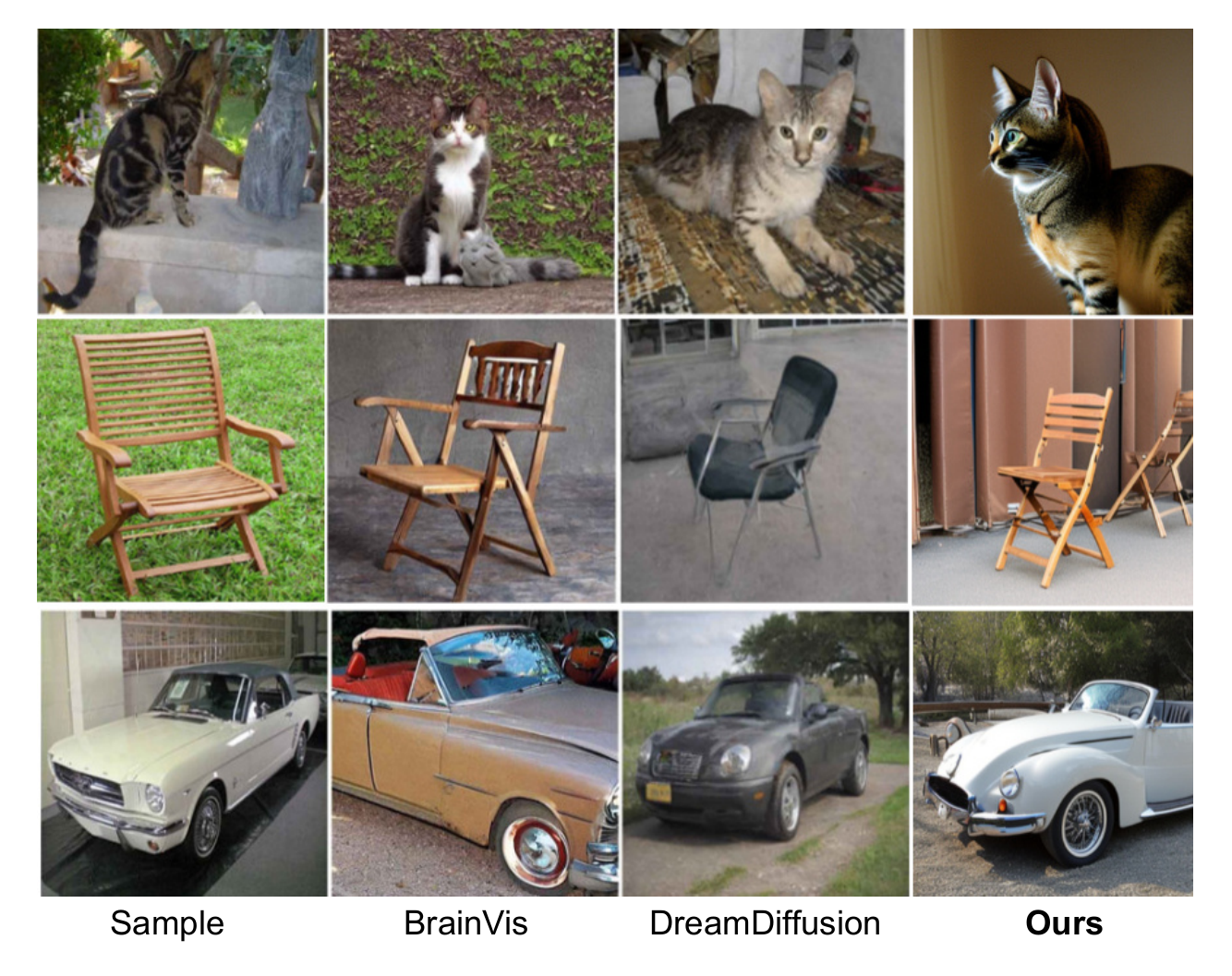}
    \caption{Comparison of images generated with models trained only on subject $4$ of EEGCVPR40. Left: random sample from the dataset.
}
    \label{fig:gt_comparision_single}
\end{figure}

\textbf{Results.}
For the EEGCVPR40 dataset, we train two variants of our method, i.e., a single and multi-subject model. For the first, the model is trained on EEGs corresponding to subject $4$ \cite{fu2023brainvis, bai2023dreamdiffusion}. Instead, the multi-subject model is trained on the whole dataset, as is done in the ThoughtViz experiments. The quantitative evaluation of these results is reported in Tab.~\ref{tab:bench_acc}. Our streamlined approach GWIT achieves state-of-the-art results in every scenario, in both generation quality and semantic correctness. 
In particular, we improve the semantic accuracy by a great margin, i.e., by $85.71$\% and $7$\% for single and multi-subject variants, respectively. This result is highly significant as this metric directly measures if the images generated from EEG signals are semantically correct with respect to the original visual cues, which is the most crucial aspect. Finally, in Fig.~\ref{fig:spampi_single} and Fig.~\ref{fig:gt_comparision_single} we present the qualitative results on EEGCPVR40, while in Fig.~\ref{fig:comparison_thoughtviz} we show results on ThoughtViz. The images generated with our method demonstrate superior quality compared to other models. Moreover, we attain semantic accuracy with respect to original visual cues as observed by comparing the generated images with samples of the original dataset in Fig.~\ref{fig:comparison_thoughtviz} and Fig.~\ref{fig:gt_comparision_single}.

\textbf{Ablations.}
In Tab.~\ref{tab:ablation_eeg} we conduct an ablation study to investigate the influence of the EEG control. Specifically, we demonstrate that images generated with EEG conditioning reach a lower LPIPS, i.e., they are closer to the ``ground truth" image, compared to images generated with only coarse control. This shows that the EEG actually guides the diffusion model to generate images semantically closer to the original visual stimuli. Lastly, in Tab.~\ref{tab:ablation_drop_guess} we investigate how the ``drop" and ``guess" modes utilized during training and inference respectively allow to prioritize the EEG conditioning. Lower LPIPS and higher accuracy indicate that the model is actually following the EEG conditioning, as they are related to distance and semantic correctness with respect to the original visual cue. We find the guess mode to be much more influential than the drop mode, however also the latter leads to an improvement.

%
\section{Conclusion}
\label{sec:conclusion}

In this paper, we have explored the use of the ControlNet adapter to handle EEG data for conditioning a latent diffusion model, allowing the reconstruction of images from EEG signals. 
We have developed a streamlined method that requires minimal preprocessing, no pretraining, and efficient fine-tuning, with the goal of moving towards real-time BCIs, while surpassing state-of-the-art methods that rely on very complex frameworks.

\balance
\bibliographystyle{IEEEtran}
\bibliography{biblio}

\end{document}